\documentclass[final]{cvpr}

\usepackage{times}
\usepackage{epsfig}
\usepackage{graphicx}
\usepackage{amsmath}
\usepackage{amssymb}
\usepackage{booktabs}
\newcommand*\samethanks[1][\value{footnote}]{\footnotemark[#1]}

\usepackage[pagebackref=true,breaklinks=true,colorlinks,bookmarks=false]{hyperref}

\def\cvprPaperID{1976} % *** Enter the CVPR Paper ID here
\def\confYear{CVPR 2021}

\begin{document}

%%%%%%%%% TITLE
\title{Learning Statistical Texture for Semantic Segmentation}

\author{
    % Authors
    Lanyun Zhu\textsuperscript{\rm1}\thanks{The first two authors contributed equally. Work done at SenseTime. Lanyun Zhu firstly proposed the paper idea.} ~~~
    Deyi Ji\textsuperscript{\rm2}\samethanks[1] ~~~
    Shiping Zhu\textsuperscript{\rm1}\samethanks[2] ~~~
    Weihao Gan\textsuperscript{\rm2}\thanks{Corresponding Author.} ~~~ 
    Wei Wu\textsuperscript{\rm2} ~~~ 
    Junjie Yan\textsuperscript{\rm2} \\
    
    Beihang University \textsuperscript{\rm1} ~~~
    SenseTime Research \textsuperscript{\rm2} \\
    {\tt\small \{zhulanyun, shiping.zhu\}@buaa.edu.cn} \\
    {\tt \small \{jideyi, ganweihao, wuwei, yanjunjie\}@sensetime.com} \\
}

% Institution1 address\\
% {\tt\small firstauthor@i1.org}
% For a paper whose authors are all at the same institution,
% omit the following lines up until the closing ``}''.
% Additional authors and addresses can be added with ``\and'',
% just like the second author.
% To save space, use either the email address or home page, not both
% \and
% Deyi Ji, Weihao Gan, Wei Wu, Junjie Yan \\
% Institution2\\
% First line of institution2 address\\
% {\tt\small secondauthor@i2.org}

% \and 

% }

\maketitle
%\thispagestyle{empty}
%%%%%%%%% ABSTRACT
\begin{abstract}
Existing semantic segmentation works mainly focus on learning the contextual information in high-level semantic features with CNNs. In order to maintain a precise boundary, low-level texture features are directly skip-connected into the deeper layers. Nevertheless, texture features are not only about local structure,  but also include global statistical knowledge of the input image. In this paper, we fully take advantages of the low-level texture features and propose a novel Statistical Texture Learning Network (STLNet) for semantic segmentation. For the first time, STLNet analyzes the distribution of low level information and efficiently utilizes them for the task. Specifically, a novel Quantization and Counting Operator (QCO) is designed to describe the texture information in a statistical manner. Based on QCO, two modules are introduced: (1) Texture Enhance Module (TEM), to capture texture-related information and enhance the texture details; (2) Pyramid Texture Feature Extraction Module (PTFEM), to effectively extract the statistical texture features from multiple scales. Through extensive experiments, we show that the proposed STLNet achieves state-of-the-art performance on three semantic segmentation benchmarks: Cityscapes, PASCAL Context and ADE20K.

\end{abstract}

%%%%%%%%% BODY TEXT
\section{Introduction}
Semantic segmentation aims to predict a label for each pixel in an image. It is one of the most fundamental 
problems in computer vision, and is widely applied in a lot of areas such as automatic driving and human-machine interaction.

Recent semantic segmentation methods mainly focus on exploiting the contextual information in high-level features with deep fully convolution based networks~\cite{fcn}. 
%High-level features are rich in context among pixels and able to address long range problems, thus achieved promising results in the past few years. 
However, only using high-level features from deep layers results in coarse and 
inaccurate output, as they are extracted from a large receptive field and miss some crucial low-level details, such as edges. To alleviate this problem, people employ skip connection to fuse low- and high-level features. DeepLabv3+~\cite{deeplabv3+} directly combines feature maps from shallow layers and deep layers before feeding them into the prediction head. Some FPN-like methods~\cite{unet,refinenet,bisenet} employ encoder-decoder structure with lateral path to refine features in a top-down manner, where multi-scale boundaries are implicitly learned. SFNet~\cite{sfnet} then applies a semantic flow to align detailed object boundaries from different levels. All of them show that low-level local features in shallow CNN layers provide the structural texture information, such as edge, which is essential for pixel-wise segmentation task.  
\begin{figure}[t]
    \centering
    \includegraphics[width=3in]{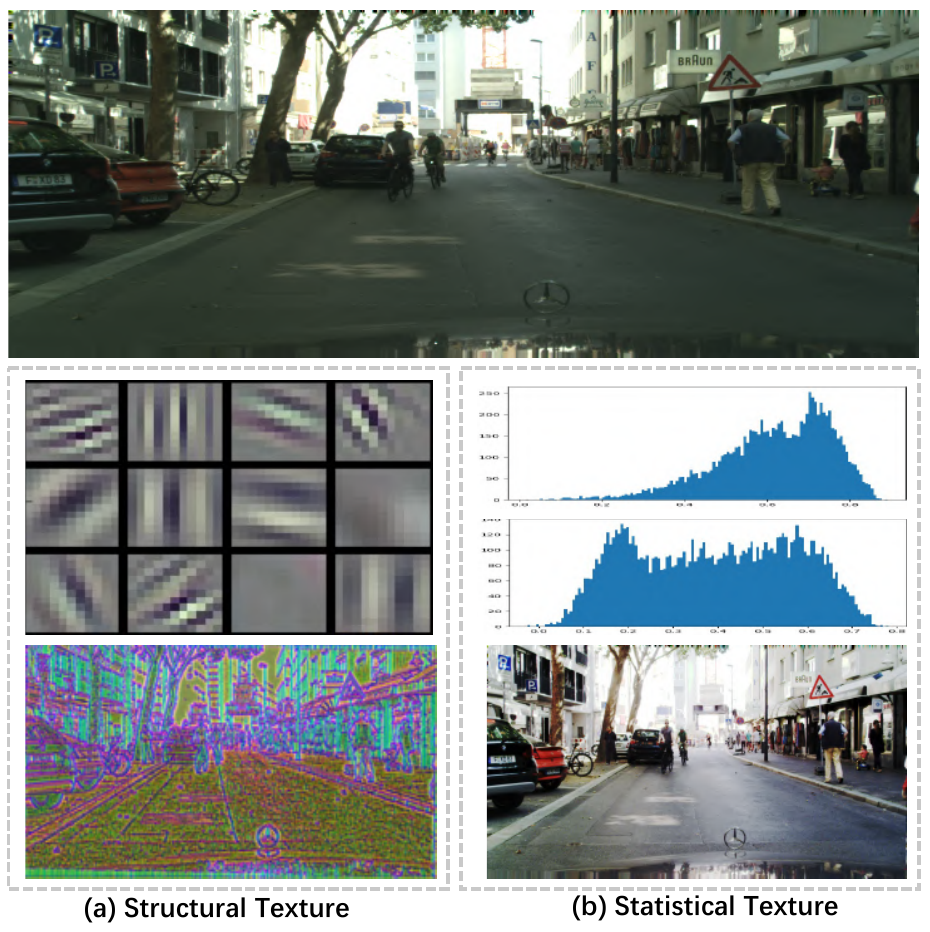}
    \caption{Examples of the structural texture and statistical texture of a image. (a) shows the convolution filters of shallow layers and the corresponding extracted structural texture in typical CNN pipeline. (b) shows the original histogram, equalized histogram and image after statistical texture enhancement, respectively.  }
    \label{texture_intro}
\end{figure}

From the perspective of digital image processing, image texture is not only about the local structural property, but also about global statistical property~\cite{gonzales2002digital, linda2001shapiro, haralick1973textural}. The structural one usually refers to some local patterns, such as boundary, smoothness and coarseness. In a typical CNN pipeline, filters in the shallow layers are good at extracting these local texture features. When visualizing those filters, we can observe that different frequencies of the local signal are analyzed (Fig. \ref{texture_intro}(a)). Therefore, structural property is also referred to spectral domain analysis~\cite{gonzales2002digital}. Another important property of texture is statistical one. Based on some low-level information (such as pixel value or local region property), it focuses on the distribution analysis of the image, such as histogram of intensity. For example in Fig. \ref{texture_intro}(b), an image captured in dark environment is usually in poor-visualization quality. After the histogram equalization enhancement step, it turns to be more detailed and better for segmentation. 
Some traditional methods try utilizing statistical texture in their tasks~\cite{arivazhagan2003texture, ramola2020study, haralick1973textural}. However, there is no mechanism in modern CNN architecture to explicitly extract and utilize the statistical texture information for semantic segmentation.  Therefore, in this paper, we propose a Statistical Texture Learning Network (STLNet) to describe and exploit statistical texture information for this task. 

Firstly, we propose a Quantization and Counting Operator (QCO) to effectively describe texture intensities in a statistical manner in deep neural networks. Specifically, statistical texture of the input image is usually of a wide variety and a continuous distribution in spectral domain, which is difficult to extract and  optimize in deep neural networks. Thus in the QCO, we first quantize the input feature into multiple levels. Each level can represent a kind of texture statistics, by which the continuous texture can be well sampled for easier description. After the quantization, the intensity of each level is then counted for texture feature encoding. 

Based on QCO, we further propose the Texture Enhancement Module (TEM) and Pyramid Texture Feature Extraction Module (PTFEM), to enhance the texture details of low-level features and exploit texture-related information from multiple scales, respectively. More comprehensively, low-level features are usually of low qualities and difficult for statistical features extraction. Inspired by histogram equalization, TEM is designed to build a graph to propagate information of all original quantization levels for texture enhancement. Furthermore, PTFEM exploits the texture information from multiple scales with 
a texture feature extraction unit and pyramid structure.

Overall, our contributions are summarized as follows:

\begin{itemize}
\item For the first time, we introduce the statistical texture information to semantic segmentation and propose a novel STLNet to take full advantage of texture information, where both low-level and high-level features are well learned in an end-to-end manner.

\item For effective description of statistical texture in deep neural networks, a novel Quantization and Counting Operator (QCO) is designed to quantize the continuous texture into multiple level intensities.

\item With the help of QCO, we propose the Texture Enhancement Module (TEM) and Pyramid Texture Feature Extraction Module (PTFEM) successively to enhance the statistical details and extract the texture features, respectively. 

\item The proposed method is practical and can be implemented in a plug-and-play fashion. Experiments show
that our method achieves state-of-the-art results on Cityscapes, Pascal Context and ADE20K datasets.
\end{itemize}

\section{Related Works}
\noindent \textbf{Semantic Segmentation. }In recent years, benefited from the rapid development
of deep learning, semantic segmentation has achieved great success. Most of the modern semantic 
segmentation models \cite{fcn, senet, unet, refinenet, reconet, acfnet, danet, hu2020class} are based on fully convolutional networks (FCN) \cite{fcn}, which first replaces the 
fully connected layers in common classification networks by convolutional layers, getting pixel-level 
prediction results.  After that, a lot of methods \cite{pspnet, deeplabv3, ann, cacnet, denseaspp, lgad, dgcwnet} are proposed to improve the basic FCN.
Some methods employ pyramid structure to capture information from different sizes of receptive fields.
PSPNet \cite{pspnet} feeds features into pyramid pooling layers with different pooling scales, and DeeplabV3 series \cite{deeplabv3,deeplabv3+} proposes an atrous spatial pyramid pooling layer consisting of multiple dilated convolutions with different dilated rates.
In our proposed methods, we also employ a pyramid structure to harvest texture information 
from different scales.\\
\noindent \textbf{Low-level Features Extraction. }
Low-level features play a crucial role in boosting performance of semantic segmentation. A lot of methods are proposed to extract and utilize low-level features. Some methods \cite{unet, refinenet, deeplabv3+, bisenet} employ skip connections to deliver low-level features to deeper layers. However, the simple multi-level features adding or concatenation operations may cause feature misalignment problems, weakening the effectiveness of low-level features. To address the issue, ANN \cite{ann} proposes a non-local block to fuse multi-level features. SFNet \cite{sfnet} aligns low-level and high-level features by learning and utilizing the semantic flow. Some works aim to exploit some more explicit low-level information such as boundary. \cite{bfp} proposes a boundary-aware feature propagation module to propagate features guided by the boundary-related low-level information. \cite{edge} explicitly generates and supervise object edges. However, all these methods only consider structural features such as edge. In contrast to them, our proposed methods further focus on statistical texture features. \\
\begin{figure*}[t]
    \centering
    \includegraphics[width=5.75in]{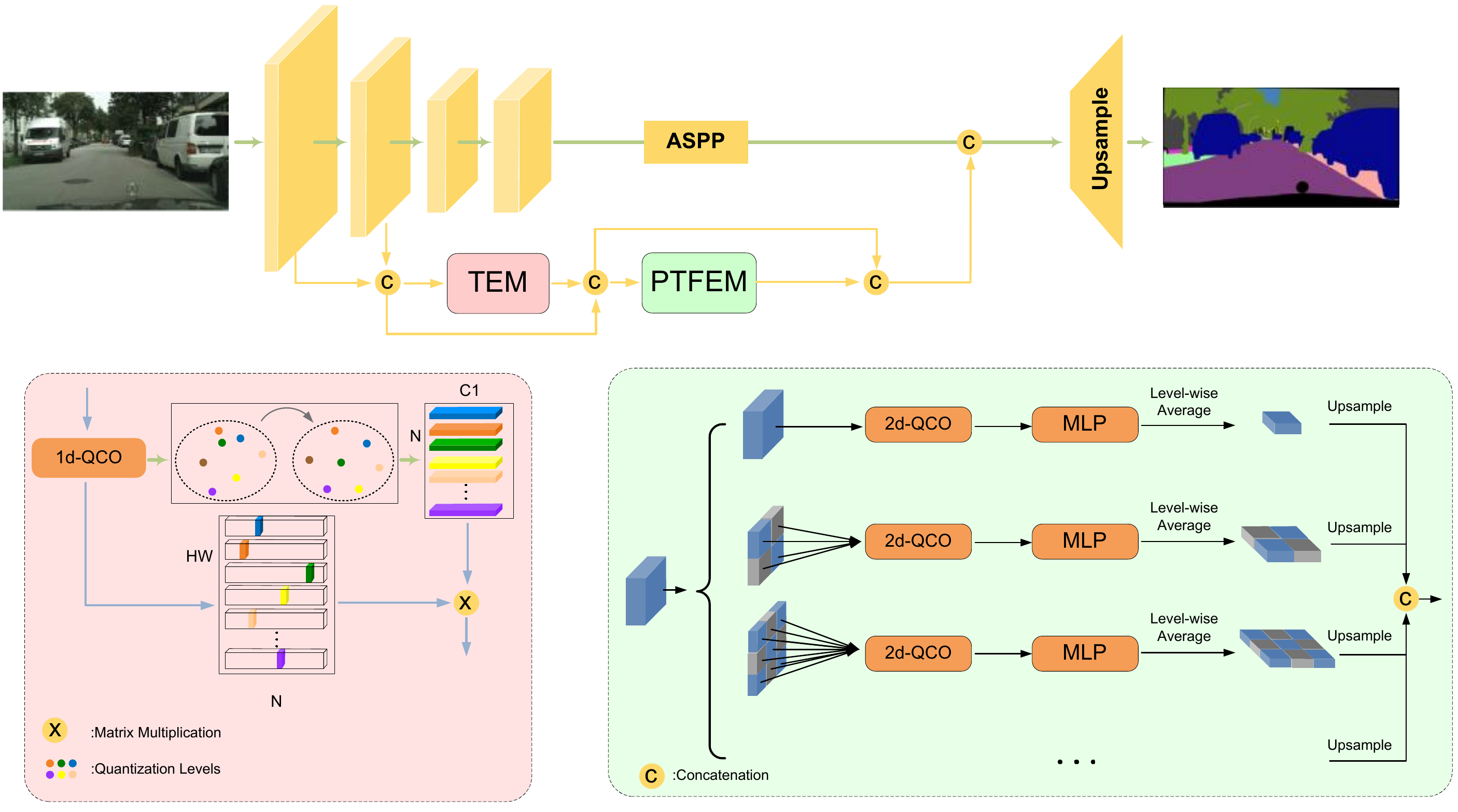}
    \caption{An overview of the proposed Statistical Texture Learning  Network (STLNet). The base network is ResNet101 followed by an ASPP module. An input image is fed into backbone ResNet-101 to extract high-level feature and multiple low-level features. The high-level feature is then fed into the ASPP module to obtain high-level context feature. We concate the first two low-level features and send into the Texture Enhance Module (TEM) to enhance the texture details, followed by the Pyramid Texture Feature Extraction Module (PTFEM) to extract the multi-scale statistical texture information. Finally, context and texture features are combined and upsampled for semantic prediction.}
    \label{overall}
\end{figure*}
\noindent \textbf{Feature Encoding}. The proposed QCO in this paper can be treated as a feature encoding method. Some previous methods \cite{deepten, contextencoding, learnable} try encoding features to better utilize information containing in them. \cite{deepten} encodes the residual vector with a learned codebook. \cite{contextencoding} employs the similar method to harvest contextual information. In contrast to them, QCO is in a self-adaptive manner when quantization, which is more stable. The most similar work may be \cite{learnable}, which encodes the feature to a learnable histogram. The main difference lies in that it quantizes each channel separately, while our methods can quantize and count the high-dimensional representation by a well-designed manner. 
%Low-level features, especially the texture features, play an important role in improving the performance of semantic segmentation.A lot of methods are proposed to extract and utilize low-level features, thus generating more accurate prediction results. Many methods \cite{unet, refinenet, deeplabv3+, bisenet} employ skip connections to deliver low-level features to deeper layers. UNet \cite{unet} connects the low-level features from downsampling encoder and high-level features from upsampling decoder. RefineNet \cite{refinenet} introduces a multi-path refinement structure to gradually deliver low-level features from shallower layers to deeper layers. However, the simple multi-level features adding or concatenation operations may cause feature alignment, weakening the effectiveness of low-level features. Therefore, some works are introduced to explore more reasonable and effective connection ways. ANN \cite{ann} proposes a non-local block to fuse multi-level features. SFNet \cite{sfnet} aligns low-level and high-level features by learning and utilizing the semantic flow. Unlike above methods, we utilize low-level features by enhancing and exploiting the texture information using statistics.

\section{Method}
In this section, we introduce the proposed Statistical Texture Learning Network (STLNet) in detail.
First, we describe the overall structure of STLNet in subsection \ref{txt_overall}. Then we expound the proposed 
Quantization and Counting Operator (QCO), Texture Enhancement Module (TEM) and Pyramid Texture Feature Extraction Module (PTFEM) in subsection \ref{txt_qcm}, 
\ref{txt_tem} and \ref{txt_tfem}, respectively. Finally, we present the loss function used to supervise the proposed network in subsection \ref{txt_loss}.

\subsection{Overall Structure} \label{txt_overall}
%In this paper, we propose a Texture Feature Enhancement and Extraction Network (TFEENet) for semantic segmentation. 
The overall structure of STLNet is shown in Fig. \ref{overall}. The network can be divided into two parts: base network and texture extraction branch. For base network, we utilize the dilated ResNet101 followed by an 
Atrous Spatial Pyramid Pooling (ASPP) module. For texture extraction branch, we first extract features from layer 1 and layer 2 of backbone ResNet, downsample them to the same size as the output of base network, and then concatenate them. After that, a Texture Enhancement Module (TEM) and Pyramid Texture Feature Extraction Module (PTFEM) are followed serially to enhance texture details and exploit texture-related information respectively. Finally, we concatenate output features from base network, TEM, PTFEM and the extracted low-level features from backbone, and pass them through a convolutional layer to get the final prediction map.

\subsection{Quantization and Counting Operator} \label{txt_qcm}
\begin{figure}[t]
    \centering
    \includegraphics[width=3.3in]{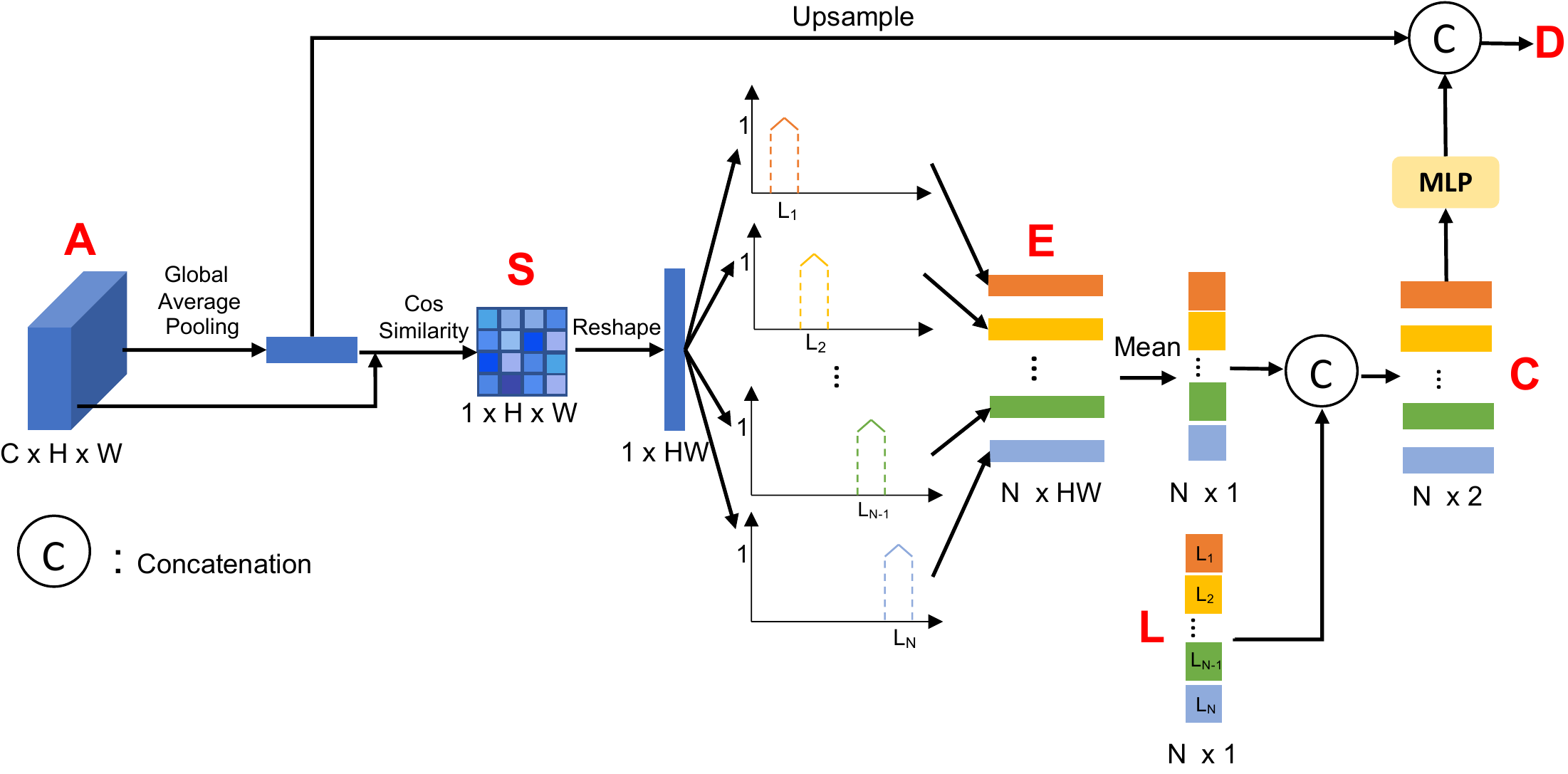}
    \caption{Structure of 1-d QCO. H and W represent feature map height and width respectively, and N represents the number of quantization levels. }
    \label{qco}
\end{figure}

Recently, most of deep networks for image processing are constructed with fully convolutional layers. Convolution operator is sensitive to the local variation, which can help to exploit some local features such as boundary. However, it is powerless for describing statistical texture. Therefore, we first propose a Quantization and Counting Operator (QCO) to describe texture in a statistical manner. Specifically, it aims to quantize the input feature into multiple levels, then count the number of features belonging to each level. QCO can be classified into two categories: One-dimensional (1-d) QCO and Two-dimensional (2-d) QCO, which are utilized in TEM and and PTFEM respectively. Both QCOs contain three parts: quantization, counting and average feature encoding. The structure of 1-d QCO is shown in Fig. \ref{qco}.

\subsubsection{1-d QCO}
\noindent \textbf{Quantization. } The input feature map of QCO is denoted as $\mathbf{A} \in \mathbb{R}^{C\times H\times W}$ and first passed through a global average pooling to get the global averaged feature $g\in \mathbb{R}^{C\times 1\times 1}$. 
For each spatial position $\mathbf{A}_{i,j}(i\in[1, W], j\in[1, H])$ on $\mathbf{A}$, we calculate the cosine similarity between $g$ and $\mathbf{A}_{i,j}$, obtaining $\mathbf{S}\in \mathbb{R}^{1\times H\times W}$. Specifically, each position $\mathbf{S}_{i,j}$ of $\mathbf{S}$ is denoted as:
\begin{equation}
    \mathbf{S}_{i,j} = \frac{g \cdot \mathbf{A}_{i,j}}{\lVert g\lVert_{2} \cdot \lVert \mathbf{A}_{i,j}\lVert_{2}}.
\end{equation}
$\mathbf{S}$ is then reshaped to $\mathbb{R}^{HW}$. Next we quantize $\mathbf{S}$ into $N$ levels $\mathbf{L} = [L_{1}, L_{2},..., L_{N}]$, and $\mathbf{L}$ is obtained by equally dividing $N$ points between the minimum and maximum values of $\mathbf{S}$. 
Concretely, the $n$th level $\mathbf{L}_{n}$ is calculated by:
\begin{equation}
    \mathbf{L}_{n} = \frac{max\left(\mathbf{S}\right) - min\left(\mathbf{S}\right)}{N} \cdot n + min(\mathbf{S}).
\end{equation}

\noindent For each spatial pixel $\mathbf{S}_{i} \in \mathbb{R}(i\in[1,HW])$, we quantize it to a quantization encoding vector $\mathbf{E}_i \in \mathbb{R}^{N}(i\in[1,HW])$, thus $\mathbf{S}$ will be finally quantized to a quantization encoding matrix $\mathbf{E} \in \mathbb{R}^{N \times HW}$. More comprehensively, $\mathbf{S}_{i}$ is quantized  with $N$ functions $F=\{f_1, f_2, ...,f_N\}$, where each $f_n$ produces $\mathbf{E}_{i,n}$ by:

% The quantization principle is, for each spatial pixel $\mathbf{S}_{i}(i\in[1,HW])$, the corresponding quantization level is the closest value in $L$. More comprehensively, We employ a mechanism similar to one-hot encoding to convert 
% each $\mathbf{S}_{i}$ to a feature vector $\mathbf{E}_{i} \in \mathbb{R}^{N}$, where the index of the maximum value in $\mathbf{E}_{i}$ can reflect the quantization level of $\mathbf{S}_{i}$. This operation can be implemented with $N$ functions, where each function $f_{n}$ maps $\mathbf{S}$ to 
% $\mathbf{E}^{n} \in \mathbb{R}^{HW}$. Specifically, for each $f_{n}$, each position $\mathbf{E}^{n}_{i}$ of output $\mathbf{E}^{n}$ is calculated by:

\begin{equation}
    \mathbf{E}_{i,n} = \left\{
        \begin{array}{lcc}
        1-|\mathbf{L}_{n}-\mathbf{S}_{i}|  & if &   -\frac{0.5}{N}\le \mathbf{L}_{n}-\mathbf{S}_{i}<\frac{0.5}{N}\\
        0 & & else
    \end{array}.
    \right.
\end{equation}
\noindent Finally, the $N$ results of $F$ are concatenated to get $\mathbf {E}_{i}$. By this way, index of the maximum value of $\mathbf{E}_{i}$ can reflect the quantization level of $\mathbf{S}_{i}$. 

Note that, in contrast to traditional argmax operation or one-hot encoding mechanism with binarization, we apply a smoother way for quantization encoding so that the  gradient vanishing problem can be avoided during the process of back propagation. 

\noindent \textbf{Counting. }
Given the quantization encoding matrix $\mathbf{E}$, we further generate the quantization counting map $\mathbf{C} \in \mathbb{R}^{N \times 2}$, where the first dimension represents each quantization level and the second dimension represents corresponding normalized counting number. Specifically, $\mathbf{C}$ is written as:
\begin{equation}
\begin{split}
    &\mathbf{C} = Cat\left(\mathbf{L}, \frac{\sum_{i=1}^{HW}{\mathbf{E}_{i,n}}}{\sum_{n=1}^{N}\sum_{i=1}^{HW}{\mathbf{E}_{i,n}}}\right),
    %,\\
    %&\mathbf{L}\in \mathbb{R}^{N}, \mathbf{L}^{n} = L_{n},
\end{split}
\end{equation}
where $Cat$ denotes concatenation operation. 

\noindent \textbf{Average Feature Encoding. }
The quantization counting map $\mathbf{C}$ reflects the relative statistics of input feature map, as the counted object is the distance from average feature. To further obtain absolute statistical information, we encode the global average feature $g$ into $\mathbf{C}$ and produce $\mathbf{D}$. We first upsample $g$ to $\mathbb{R}^{N\times C}$, and then obtain $\mathbf{D}$ by:
\begin{equation}
    \begin{split}
        \mathbf{D} = Cat\left(MLP\left(\mathbf{C}\right), g\right),
    \end{split}
\end{equation}
where $MLP$ aims to increase the dimension for $\mathbf{C}$. It contains two layers, in which the first one is followed by a Leaky ReLU to enhance nonlinearity. 

Both $\mathbf{E}$ and $\mathbf{D}$ will be output, and denote the quantization encoding map and statistical feature respectively. 

\subsubsection{2-d QCO } \label{2-d QCO}
The output of 1-d QCO reflects the distribution of features from all spatial positions. However, it carries no information regarding spatial
relationships between pixels, which plays an important role in describing texture. To this end, we further propose a 2-d QCO to count the distribution of co-occurring pixel features. 

\noindent \textbf{Quantization.} 
Quantization in 2-d QCO is intent to count the co-occurring spatial relationships among pixels in the input feature map, and can be extended by that in 1-d QCO. 
Specifically, the input $\mathbf{A}\in \mathbb{R}^{C\times H\times W}$ is first passed through the similar procedure as 1-d QCO to get quantization encoding map $\mathbf{E}$ and quantization levels $\mathbf{L}$. We then reshape $\mathbf{E}$ to $\mathbb{R}^{N\times 1\times H\times W}$.
%Specifically, for input $\mathbf{A}\in \mathbb{R}^{C\times H\times W}$, given the quantization encoding map $\mathbf{E}\in \mathbb{R}^{N\times HW}$ obtained by the similar procedure in 1-d QCO, we first reshape it to $\mathbb{R}^{N\times H\times W}$. 
For each pair of adjacent pixels $ \textbf{E}_{i,j} \in \mathbb{R}^{N \times 1}$ and $ \textbf{E}_{i,j+1} \in \mathbb{R}^{N \times 1}$, we calculate their product $\mathbf{\hat{E}}_{i, j}$:

\begin{equation}
\label{2dQCO}
    \mathbf{\hat{E}}_{i, j} = \textbf{E}_{i,j} \cdot \textbf{E}_{i,j+1}^T,
\end{equation}

% \begin{equation}
%     \mathbf{\hat{E}}_{h, w} = \mathbf{E}_{h, w} \otimes \left(\mathbf{E}_{h, w+1}\right)^{T},
% \end{equation}

% \begin{equation}
%     \mathbf{\hat{E}} = \mathbf{E} \cdot \mathbf{E}^T,
% \end{equation}

\noindent where $T$ refers to matrix transposition. Only when $\mathbf{A}_{i, j}$ is quantized to $\mathbf{L}_m$ and $\mathbf{A}_{i, j+1}$ is quantized to $\mathbf{L}_n$, the corresponding $\mathbf{\hat{E}}_{m, n, i, j}$ dose not equal to zero.  Therefore, $\mathbf{\hat{E}} \in \mathbb{R}^{N\times N\times H\times W}$ can represent the quantization co-occurrence of every two adjacent pixels.  

% Note that, we set $\mathbf{E}_{:, W+1}$ to a zero vector for the convenience of calculation. Only when the values of adjacent pixels $\mathbf{E}_{h,w}$ and $\mathbf{E}_{h, w+1}$ form a 2-d vector $[m, n]$, the corresponding $\mathbf{\hat{E}}_{m, n, h, w}$ dose not equal to zero. 

\noindent \textbf{Counting. }
Given  $\mathbf{\hat{E}}$, we generate a 3-dimensional map $\mathbf{C}$, where the first two dimensions represent each possible quantization co-occurrence and the third dimension represents the corresponding normalized counting number. $\mathbf{C}$ is denoted as:
\begin{equation}
\begin{split}
    &\mathbf{C} = Cat\left(\hat{\mathbf{L}}, \frac{\sum_{i=1}^{H}\sum_{j=1}^{W}{\mathbf{\hat{E}}_{m, n, i, j}}}{\sum_{m=1}^{N}\sum_{n=1}^{N}\sum_{i=1}^{H}\sum_{j=1}^{W}{\mathbf{\hat{E}}_{m, n, i, j}}}\right),\\
    &\hat{\mathbf{L}} \in \mathbb{R}^{2\times N\times N},\quad \hat{\mathbf{L}}_{m, n} = [\mathbf{L}_{m}, \mathbf{L}_{n}],
\end{split}
\end{equation}
where $\hat{\mathbf{L}}$ denotes all possible quantization levels pairs of adjacent pixels.

\noindent \textbf{Average Feature Encoding. }Following 1-d QCO, we also encode the average feature to obtain absolute representation. Denote the average feature of processed region as $g\in \mathbb{R}^{C}$, we upsample it to $\mathbb{R}^{N\times N\times C}$. Then the final output $\mathbf{D}$ is obtained by:
\begin{equation}
    \mathbf{D} = Cat\left(MLP(\mathbf{C}), g\right).
\end{equation}

\subsection{Texture Enhancement Module} \label{txt_tem}
Low-level features extracted from shallow layers of backbone network are often with low quality, especially with low contrast, causing the texture details 
to be ambiguous, which has negative impacts on the extraction and utilization of low-level information. Therefore, we propose a Texture Enhancement Module (TEM) to 
enhance the texture details of low-level features so that it is easier to capture texture-related information in the following steps.

The way we enhance texture is inspired by histogram quantification, a classical method for image quality enhancement. Specifically, this method first produces a histogram, whose horizontal axis and vertical axis represent each gray level and its counting value respectively. We denote these two axes as two feature vectors $\mathbf{G}$ and $\mathbf{F}$. Histogram quantification aims to reconstruct levels to $\mathbf{G}'$ using the statistical information containming in $\mathbf{F}$. Each level 
$\mathbf{G}_{n}$ is transformed to $\mathbf{G}'_{n}$ by: 
\begin{equation} \label{gcn}
    \mathbf{G}'_{n} = \frac{(N-1)\sum_{i=0}^{n}\mathbf{F}_{n}}{\sum_{i=0}^{N}\mathbf{F}_{n}},
\end{equation}
where $N$ is the total number of gray levels.

In the proposed TEM, we first pass the input feature map through a 1-d QCO to get the quantization encoding map $\mathbf{E}\in \mathbb{R}^{N\times HW}$ and statistics feature $\mathbf{D}\in \mathbb{R}^{C_{1}\times N}$, where $\mathbf{D}$ plays the role of histogram. After that, we get the new quantization levels $\mathbf{L}'$ using $\mathbf{D}$. Inspired by Equation. \ref{gcn}, each new level should be obtained by perceiving statistical information of all original levels, which can be treated as a graph. To this end, 
%Utilizing the statistical information containing in $\mathbf{E}$ reasonably is of great importance, and our proposed ways are inspired by histogram quantification,
%After that, we reconstruct each quantification level by modeling the relationship among statistical features of all levels using a non-local operation. Specifically, the
%calculation process is as follows:
%a classical method for image quality enhancement. Denote the histogram as two feature vectors $\mathbf{G}$ and 
%$\mathbf{F}$ representing the gray levels and their corresponding counting values respectively. Histogram quantification aims to reconstruct levels to $\mathbf{G}'$ using the statistical information $\mathbf{F}$. Each level 
%$\mathbf{G}_{n}$ is transformed to $\mathbf{G}'_{n}$ by:
%\begin{equation}
 %   \mathbf{G}'_{n} = %\frac{(N-1)\sum_{i=0}^{n}\mathbf{F}_{n}}{\sum_{i=0}^{N}\mathbf{F}%_{n}}
%\end{equation}
%where $N$ is the total number of gray levels. From this formula we can find that the core idea of level transformation lies in analysising from a global view by perceiving statistical information of all gray levels. 
we build a graph to propagate information from all levels. The statistical 
feature of each quantization level is defined as a node. In the traditional histogram quantification algorithm, the adjacent matrix is a manually defined diagonal matrix, and we extend it to a learned one as follows:
\begin{equation}
    \mathbf{X} = Softmax\left(\phi_{1}\left(\mathbf{D}\right)^{T} \cdot \phi_{2}\left(\mathbf{D}\right)\right),
\end{equation}
where $\phi_{1}$ and $\phi_{2}$ refer to two different $1\times 1$ convolutions, and Softmax performed on the first dimension works as a nonlinear normalization function.
Then we update each node by fusing features from all other nodes, getting the reconstructed quantization levels $\mathbf{L}'\in \mathbb{R}^{C_{2}\times N}$:
\begin{equation}
    \mathbf{L}' = \phi_{3}\left(\mathbf{D}\right) \cdot \mathbf{X},
\end{equation}
where $\phi_{3}$ refers to another $1\times 1$ convolution. 

After that, we assign the reconstructed levels $\mathbf{L}'$ to each pixel to get the final output $\mathbf{R}$ using the quantization encoding map $\mathbf{E}\in \mathbb{R}^{N\times HW}$, since $\mathbf{E}$ can reflect the original quantization level of each pixel. $\mathbf{R}$ is obtained by:
\begin{equation}
    \mathbf{R} = \mathbf{L}' \cdot \mathbf{E}.
\end{equation}
Finally, $\mathbf{R}$ is reshaped to $\mathbf{R}^{C_{2}\times H\times W}$.

\subsection{Pyramid Texture Feature Extraction Module} \label{txt_tfem}
We further propose a Pyramid Texture Feature Extraction Module (PTFEM), which aims to exploit texture-related information from multiple scales using feature maps containing rich texture details. We first describe the unit to capture texture features from each processed region, then introduce the pyramid structure to build PTFEM.

\noindent \textbf{Texture Feature Extraction Unit. } Texture is highly correlated with the statistical information about spatial relationships between pixels. The way we extract texture information is inspired by gray level co-occurrence matrix (GLCM), a widely used method to capture texture representations. GLCM first produces a co-occurrence matrix $\mathbf{M}\in \mathbb{R}^{H\times W}$ where the value $\mathbf{M}_{h, w}$ of each position on $\mathbf{M}$ denotes the number of occasions that the gray values of two adjacent pixels form a 2-d vector $[h, w]$. Then some statistical descriptors such as contrast, uniformity and homogeneity are employed to represent the texture information of the region. In our proposed methods, for the processed region of feature map, we first feed it into the 2-d QCO to get the co-occurrence statistics feature $\mathbf{F}\in \mathbb{R}^{C\times N \times N}$, where $C$ denotes the channel number and $N$ denotes the number of quantization levels. 
Different from the hand-crafted descriptors used in GLCM, benefited from the powerful feature extraction ability of deep learning, we force the network to learn useful statistical representations automatically. Specifically, an MLP followed by a level-wise average is employed to produce the texture feature $\mathbf{T}$ of processed region: 
\begin{equation}
\begin{split}
        &\mathbf{F}' = MLP\left(\mathbf{F}\right),\quad \mathbf{F}'\in \mathbb{R}^{C'\times N\times N},\\
        &\mathbf{T} = \frac{\sum_{m=1}^{N}\sum_{n=1}^{N}{\mathbf{F}'_{:, m, n}}}{N \cdot N}.
\end{split}
\end{equation}
\noindent \textbf{Pyramid Structure. }As discovered by some previous works \cite{pspnet, deeplabv3}, multi-scale features help to boost the performance and robustness of semantic segmentation effectively. Such features can be captured by the pyramid structure such as Spatial Pyramid Pooling and Atrous Spatial Pyramid Pooling. Inspired by the success of these methods, we also employ a pyramid structure to carve texture features from multiple scales. Specifically, the pyramid structure passes input feature map through four parallel branches with different scales [1,2,4,8]. For each branch, the feature map is separated into different number of sub-regions, and each sub-region is passed through the texture feature extraction unit to exploit the region's corresponding texture representation. Then we upsample the obtained texture feature map of each branch to the original size as input map via nearest interpolation, and concatenate the output of four branches, getting the output of PTFEM.

\subsection{Loss} \label{txt_loss}
Following \cite{pspnet}, we apply deep supervision to make the deep network easier to train.
Specifically, besides getting the prediction map from the final output, we also add an auxiliary layer 
after the backbone ResNet layer 3, getting an auxiliary prediction map. Both prediction maps are supervised.
The overall loss can be denoted as:
\begin{equation}
    \mathcal{L} = \mathcal{L}_{f} + \alpha.\mathcal{L}_{a},
\end{equation}
where $\alpha=0.4$ is a hyperparameter to balance the weight between final prediction loss $\mathcal{L}_{f}$
and auxiliary prediction loss $\mathcal{L}_{a}$. For $\mathcal{L}_{a}$, we use the common cross entropy loss. 
For $\mathcal{L}_{f}$, we employ the online hard examples mining (OHEM) loss to handle the problem caused by 
class imbalance. Hard examples refer to pixels whose predicted probabilities of their corresponding correct classes are smaller
than $\theta$. To ensure a stable training process, we keep at least $K$ pixels within each batch for training. In our experiments,
$\theta$ and $K$ are set to 0.7 and 100000, respectively.

\section{Experiments}
\subsection{Datastes}
We evaluate our methods on three popular datasets, including Cityscapes, PASCAL Context and ADE20K.
\textbf{Cityscapes.} Cityscapes is a large urban scene understanding dataset. It contains 5000 finely annotated images and 20000 coarsely annotated images. We only use the 5000 finely annotated images for experiments, which can be divided into 2975/500/1525 images for training, validation and testing, respectively. %Following\cite{deeplabv3}, only 19 classes among the total 30 classes are utilized for evaluation. \\

\noindent \textbf{PASCAL Context. }PASCAl Context is an extended dataset for PASCAL 2010 by providing more detailed segmentation annotation. It contains 4998 images for training and 5105 images for validation. 
%We use 59 semantic classes along with 1 background class for evaluation.\\

\noindent \textbf{ADE20K. }ADE20K is a large and challenging dataset for scene parsing. It is consisted of 30K, 10K and 10K images for training, validation and testing, respectively. 150 classes are utilized for evaluation.

\subsection{Implementation Details}
The ResNet101 pretrained on ImageNet is used as the backbone of the propose STLNet. Following \cite{pspnet}, we replace the last two downsampling operations by dilated convolutional layers with dilation rates being 2 and 4 respectively, making the output stride to 8. All BN layers in the network are replaced by Sync-BN. Quantization level is 128 in 1-d QCO and 8 in 2-d QCO. We apply stochastic gradient descent(SGD) as the optimizer. For training, we use `poly' policy to set the learning rate, where the learning rate for each iteration equals to initial rate multiplied by $(1-\frac{iter}{max\_iter})^{0.9}$. 
We apply random scaling in the range of [0.5, 2], random cropping and random left-right flipping for data augmentation. For Cityscapes, we set the initial learning rate as 0.01, weight decay as 0.0005, crop size as $769\times 769$, batch size as 8 and training iteration as 40K. For PASCAL Context. we set the initial learning rate as 0.001, weight decay as 0.0001, crop size as $513\times 513$, batch size as 16 and training iterations as 30K. For ADE20K, we set the initial learning rate as 0.01, weight decay as 0.0005, crop size as $513\times 513$, batch size as 16 and training iteration as 100K. All experiments are conducted using 8 $\times$ NVIDIA TITAN Xp GPUs.
\subsection{Ablation Study}
All the following ablation study experiments are conducted on Cityscapes.

\begin{table}[t]
    \centering
    \begin{tabular}{l c}
       \toprule
       Method & mIoU(\%)\\
       \midrule
       ResNet-101  &  76.4\\
       ResNet-101 + SLF & 77.0\\
       ResNet-101 + SLF + TEM  & 80.3\\
       ResNet-101 + SLF + PTFEM & 80.0\\
       ResNet-101 + SLF + TEM + PTFEM & 80.9\\
       \midrule
       ResNet-101 + ASPP&  78.8\\
       ResNet-101 + ASPP + SLF & 79.3\\
       ResNet-101 + ASPP + SLF + TEM  & 80.9\\
       ResNet-101 + ASPP + SLF + PTFEM & 80.5\\
       ResNet-101 + ASPP + SLF + TEM + PTFEM & 81.5\\
       \bottomrule
    \end{tabular}
    \caption{Ablation studies of different components in STLNet. SLF refers to low-level features (SLF) fusion.}
    \label{ablation1}
\end{table}

\noindent \textbf{Ablation of STLNet.}
We conduct experiments to verify the effectiveness of different components in STLNet. Experimental results are shown in Table \ref{ablation1}. We first choose ResNet101 with dilated convolutions as the baseline, which achieves 76.4\% mIoU. First, we simply concatenate features from backbone ResNet layer1 and layer2, denoted as shallow layer features (SLF), and then connect them with the output of base network. This modification slightly improves the performance to 77.0\%. Then we plug TEM and PTFEM after SLF individually, increasing mIoU to 80.3\% and 80.0\% respectively.
%Due to TEM and TFEM modules are plugged after low-level features extracted from layer1 and layer2 of backbone ResNet, we first evaluate the performance improvement brought by original low-level features. Concretely, we concate features from backbone layer2 with output features of backbone layer4, then feed theminto a classification head to get the prediction results. This modification slightly improves the performance to xx.xx. 
Finally, by employing both TEM and PTFEM, the mIoU result is improved to 80.9\%, which is 4.5\% better than the baseline.

We further perform experiments based on a more powerful network ResNet101 + ASPP and repeat experiments mentioned above. 
TEM and PTFEM also improve the performance significantly in this case, and finally achieves 81.5\% mIoU under ResNet101 + ASPP + SLF + TEM + PTFEM setting. It demonstrates that TEM and PTFEM can be generally plugged into any existing semantic segmentation network to improve performance. 

\begin{figure}[t]
    \centering
    \includegraphics[width=3.2in]{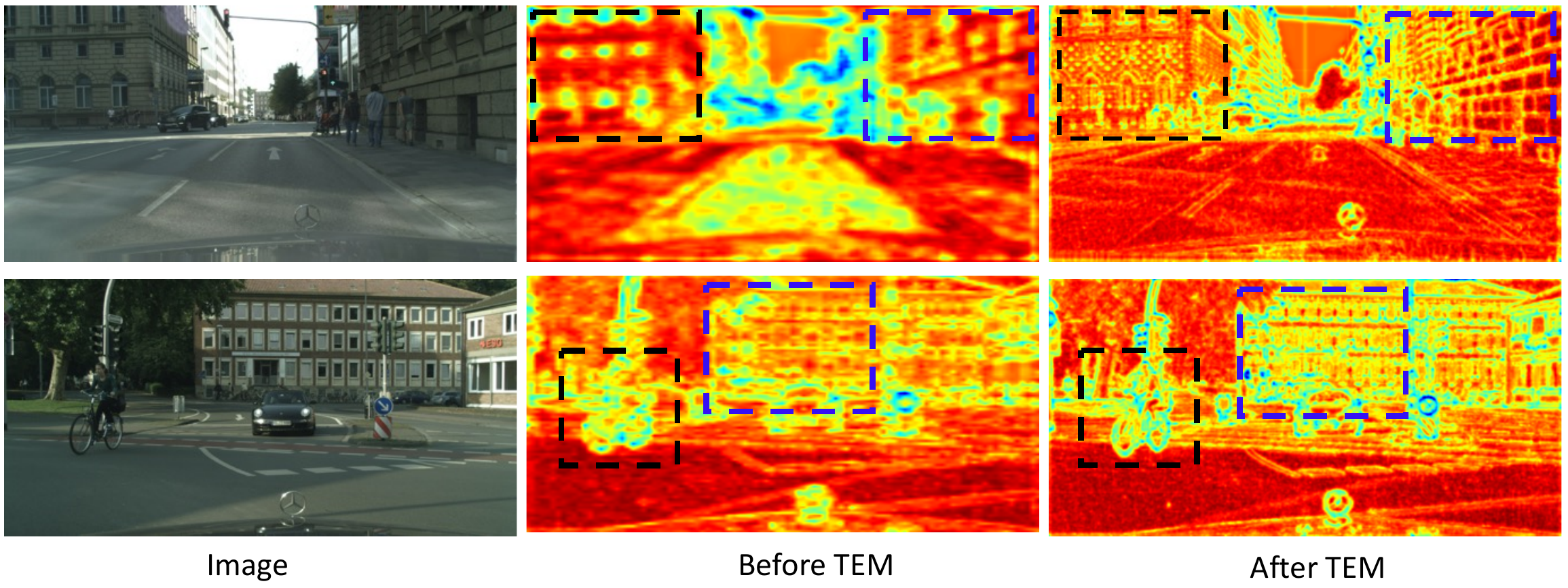}
    \caption{Visualization comparison on feature maps before and after TEM. In the indicated regions (marked by the black and blue boxes), the texture details of buildings and bicycle is clearly enhanced after TEM.}
    \label{visual_tem}
\end{figure}

\begin{figure}[t]
    \centering
    \includegraphics[width=2.3in]{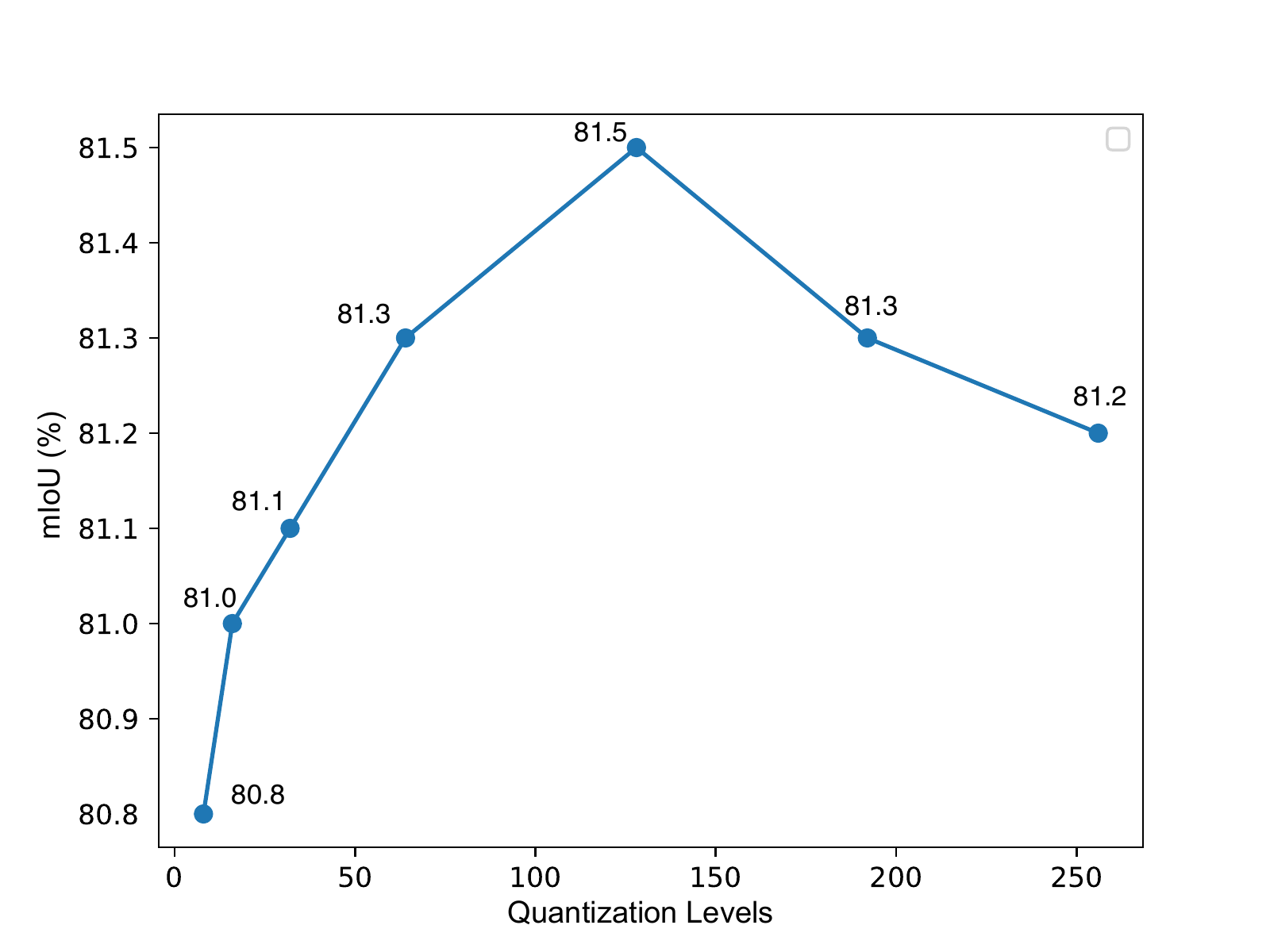}
    \caption{Ablation of quantization levels. The horizontal axis refers to number of quantization leves adopted by 1-d QCO, and vertical axis refers to the corresponding mIoU. }
    \label{n}
\end{figure}

\begin{table}[t]
    \centering
    \begin{tabular}{l c}
       \toprule
       Method & mIoU(\%)\\
       \midrule
       ResNet-101 + ASPP + SLF  &  79.3\\
       \midrule
       ResNet-101 + ASPP + SLF + TEM(w/o graph) & 79.9\\
       ResNet-101 + ASPP + SLF + TEM(w/ graph) & 80.9\\
       \bottomrule
    \end{tabular}
    \caption{Ablation studies of the graph in TEM.}
    \label{ablation_graph}
\end{table}

\noindent \textbf{Abaltion of Quantization Levels.}
During the quantization process, the number of quantization levels can be chosen flexibly, and we conduct experiments to explore the most suitable choice and results are shown in Fig. \ref{n}. Specifically, we use ResNet-101 + ASPP + SLE + TEM + PTFEM as the baseline, and report results of 6 different numbers (8, 16, 32, 64, 128, 256) of quantification levels used in 1-d QCO in TEM. The performance is unsatisfactory when the number of quantization levels is too small, and increasing the number to 128 can significantly boost mIoU to 81.5\%.  However, further increasing the number may cause the performance to decrease slightly. The reason may be that a too dense quantization will result in a sparse high-dimensional statistics feature, which may affect the training stability negatively and cause overfitting.

%\noindent \textbf{Ablation of Shallow Layer Features. }The TE module and PTFE module are plugged after features extracted from shallow layers of the base network. We find that the position where the processed shallow layer features are extracted affect the performance to some extent. 
\noindent \textbf{Ablation of Graph in TEM.}
The proposed TEM can be dived into three parts: 1-d QCO, graph construction and feature reassignment, and we evaluate the impact of the graph in TEM. We use ResNet-101 + ASPP + SLF with 79.3\% mIoU as the baseline. As shown in Table \ref{ablation_graph}, we first plug 1-d QCO on the baseline, and directly assign features obtained from it to each pixel, getting 79.9\% mIoU. Further building the graph boosts mIoU to 80.9\%. It demonstrates the importance of getting reconstructed quantization levels by perceiving all original levels.  
%\noindent \textbf{Ablation of TEM. }
%In this part, we design multiple experiments to investigate the effectiveness of different components and settings in TEM. We choose ResNet101 + ASPP + SLF + PTFEM with 80.5\% mIoU as baseline, and adding TE module with different settings between SLF and PTFE module for ablation studies. The results are shown in Table \ref{ablation2}. TEM can be divided into three parts: 1-d QCO, graph construction and feature assignment. We first plug 1-d QCO on the baseline, and directly assign features obtained from it to each pixel, getting 80.8\% mIoU. Further building the graph boosts mIoU to 81.5\%, which verifies the importance of reconstructing levels from a global view. During the quantization operation, the number of quantization levels can be chosen flexibly, and we conduct experiments to explore the most suitable choice. Specifically, we report results of 6 different numbers of quantification levels: 8, 16, 32, 64, 128, 256. The performance improvement is limited when the number of quantification level is too small, and increasing the number to 128 can significantly boost mIoU to 81.5\%. However, further increasing the number may cause the performance to decrease slightly. The reason may be that a too dense quantization will result in a sparse high-dimensional statistics feature, which may affect the training stability negatively and cause overfitting.We also provide visualization of feature maps before and after TEM, which is presented in Fig. \ref{visual_tem}. It shows that TEM refine and enhance the texture details significantly.

\begin{table}[t]
    \centering
    \begin{tabular}{c c c c c}
       \toprule
       1 & 2 & 3 & 6 & mIoU(\%)\\
       \midrule
       \checkmark & & & & 81.0\\
       & \checkmark & &  & 81.0 \\
       & & \checkmark & & 81.1\\
       & & & \checkmark & 81.2\\
       \checkmark & \checkmark & \checkmark & \checkmark &  81.5\\
       \bottomrule
    \end{tabular}
    \caption{Ablation of different pyramid scales adopted by PTFEM.}
    \label{ablation_scale}
\end{table}

\noindent \textbf{Ablation of Pyramid Scales. } The proposed PTFEM is constructed in a pyramid manner to harvest texture information from multiple scales. To verify the effectiveness of multi-scale features, we evaluate performance for models with different scale rates adopted in PTFEM. As shown in Table \ref{ablation_scale}, extracting multi-scale texture features outperforms single-scale features by a large margin, which verifies the necessity of employing pyramid structure. 
%\noindent \textbf{Ablation of PTFEM.} We further perform experiments to evaluate the impact of PTFEM. We use ResNet101 + ASPP + SLF + TEM as the baseline. PTFEM is constructed in a pyramid manner to capture multi-scale texture information, whose structure is similar to pyramid spatial pooling (PSP) module. we first directly plug a PSP module after the TEM and mIoU is not increased. Then we replace PSP module by PTFEM and a significant performance improvement (+0.6\%) is achieved, demonstrating the effectiveness of extracting texture features via proposed methods. We also evaluate whether the multi-scale features help boost performance compared with single-scale feature. Specifically, we report results of different scale rates adopted in PTFEM. It shows that extracting multi-scale texture features outperform single-scale features by a large margin, verifying the necessity of employing pyramid structure. \\
\begin{table}[t]
    \centering
    \begin{tabular}{l c c c}
       \toprule
       Method & Flip & MS & mIoU(\%)\\
       \midrule
       STLNet & & & 81.5\\
       STLNet & \checkmark & & 81.7 \\
       STLNet & & \checkmark & 81.9 \\
       STLNet & \checkmark & \checkmark & 82.3 \\
       \bottomrule
    \end{tabular}
    \caption{Ablation studies of evaluation strategies.}
    \label{strategies}
\end{table}

\noindent \textbf{Ablation of Evaluation Strategies. }
Following \cite{ccnet, danet}, we apply some strategies for evaluation, including left-right flipping and multi-scale input with scale sizes [0.75, 1, 1.25, 1.5, 1.75, 2]. We conduct experiments based on ResNet101 + ASPP + SLF + TE + PTFE. As shown in Table \ref{strategies}, Flip and MS improve performance by 0.2\% and 0.4\% respectively, and adopting both strategies boost mIoU to 82.3\%.

\begin{table}[t]
    \centering
    \scalebox{0.9}{
    \begin{tabular}{l c c}
       \toprule
       Method & FLOPS & Parameters\\
       \midrule
       Baseline & 275.55G & 80.08M\\
       \midrule
       TEM & 15.76G & 0.94M \\
       PTFEM & 1.72G & 0.37M \\
       TEM + PTFEM & 17.48G & 1.31M\\
       \bottomrule
    \end{tabular}
    }
    \caption{Flops and parameters.}
    \label{flop}
\end{table}
\noindent \textbf{FLOPs and Parameters. } We report the FLOPs and parameters of proposed methods in Table \ref{flop}, which are estimated for an input image with size of $1\times 3\times 769\times 769$. It shows that TEM and PTFEM are lightweight and only bring very little extra cost.

\noindent \textbf{Visualization of Features. }TEM is proposed to enhance the texture details of low-level features. To verify the effectiveness intuitively, we visualize and compare feature maps before and after TEM, which is shown in Fig. \ref{visual_tem}. After passing through TEM, the feature maps are refined significantly and texture details become more clear. 

\begin{table}[t]
    \centering
    \scalebox{0.9}{
    \begin{tabular}{l c c}
       \toprule
       Method & Backbone & mIoU(\%)\\
       \midrule
       PSANet \cite{psanet}& ResNet-101 & 80.1\\
       ANN \cite{ann}& ResNet-101 & 81.3\\
       CPNet \cite{cpnet}& ResNet-101 & 81.3\\
       CCNet \cite{ccnet}& ResNet-101 & 81.4\\
       DANet \cite{danet}& ResNet-101 & 81.5\\
       SpyGR \cite{spygr}& ResNet-101 & 81.6 \\
       ACFNet \cite{acfnet}& ResNet-101 & 81.8\\
       OCR \cite{ocr}& ResNet-101 & 81.8\\
       SFNet \cite{sfnet}& ResNet-101 & 81.8\\
       \midrule
       Ours & ResNet-101 & \textbf{82.3}\\
       \bottomrule
    \end{tabular}
    }
    \caption{Results comparison on Cityscapes test set.}
    \label{cityscapes}
\end{table}

\begin{figure}[t]
    \centering
    \includegraphics[width=3.3in]{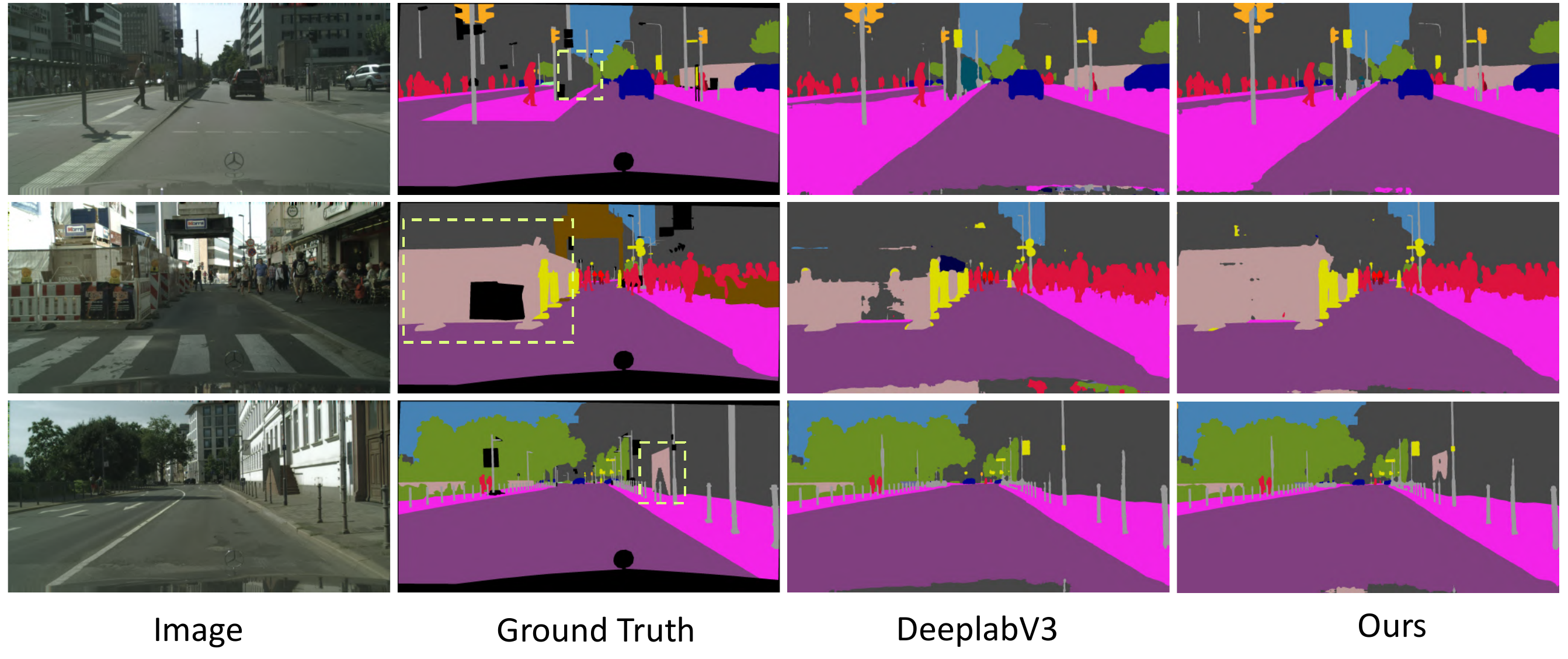}
    \caption{Visualization comparison on Cityscapes validation set.}
    \label{visual_comp}
\end{figure}

\subsection{Comparison with State-of-the-arts}
We further report performance comparison with state-of-the-art methods on three datasets: Cityscapes, PASCAL Context and ADE20K. For Cityscapes, we train the proposed STLNet using both training set and validation set, and evaluate on the test set. For another two datasets, we train model on the training set and report results on the validation set. Results on three datasets are shown in Table \ref{cityscapes}, Table \ref{pascal_context} and Table \ref{ade20k}, respectively, and it can be observed that STLNet achieves state-of-the-art performance on all datasets. We also provide visual comparison results with DeeplabV3 on Cityscapes validation set in Fig. \ref{visual_comp}.

\begin{table}[t]
    \centering
    \scalebox{0.94}{
    \begin{tabular}{l c c}
       \toprule
       Method & Backbone & mIoU(\%)\\
       \midrule
       FCN-8s \cite{fcn}& & 37.8 \\
       VeryDeep \cite{verydeep} & & 44.5\\
       Deeplab-v2 \cite{deeplabv2}& ResNet-101 & 45.7\\
       RefineNet \cite{refinenet} & ResNet-152 & 47.3\\
       PSPNet \cite{pspnet}& ResNet-101 & 47.8\\
       MSCI \cite{msci}& ResNet-152 & 50.3 \\
       EncNet \cite{encnet}& ResNet-101 & 51.7 \\
       DANet \cite{danet}& ResNet-101 & 52.6 \\
       SpyGR \cite{spygr}& ResNet-101 & 52.8\\
       SVCNet \cite{svcnet}& ResNet-101 & 53.2\\
       CPNet \cite{cpnet}& ResNet-101 & 53.9\\
       CFNet \cite{cfnet}& ResNet-101 & 54.0\\
       DMNet \cite{dmnet}& ResNet-101 & 54.4\\
       RecoNet \cite{reconet}& ResNet-101 & 54.8\\
       CaC-Net \cite{cacnet}& ResNet-101 & 55.4\\
       \midrule
       Ours & ResNet-101 & \textbf{55.8}\\
       \bottomrule
    \end{tabular}
    }
    \caption{Results comparison on PASCAL Context validation set.}
    \label{pascal_context}
\end{table}

\begin{table}[t]
    \centering
    \scalebox{0.94}{
    \begin{tabular}{l c c}
       \toprule
       Method & Backbone & mIoU(\%)\\
       \midrule
       RefineNet \cite{refinenet}& ResNet-152 & 40.70 \\
       PSPNet \cite{pspnet}& ResNet-101 & 43.29\\
       SAC \cite{pspnet}& ResNet-101 & 44.30\\
       EncNet \cite{encnet}& ResNet-101 & 44.65\\
       CFNet \cite{cfnet}& ResNet-101 & 44.89\\
       CCNet \cite{ccnet}& ResNet-101 & 45.22\\
       RecoNet \cite{reconet}& ResNet-101 & 45.54\\
       CaC-Net \cite{cacnet}& ResNet-101 & 46.12\\
       CPNet \cite{cpnet}& ResNet-101 & 46.27\\
       \midrule
       Ours & ResNet-101 & \textbf{46.48}\\
       \bottomrule
    \end{tabular}
    }
    \caption{Results comparison on ADE20K validation set.}
    \label{ade20k}
\end{table}

\section{Conclusion}
In this paper, we propose a novel STLNet for semantic segmentation. The key innovation lies in learning texture features from the statistical perspective. Specifically, we propose a Quantization and Counting Operator to describe statistical texture representations, and then use Texture Enhancement Module and Pyramid Texture Feature Extraction Module to enhance details and exploit texture-related low-level information, respectively. Extensive ablation experiments demonstrate the effectiveness of proposed methods. And comparison with state-of-the-art methods shows that STLNet achieves outstanding performance on multiple datasets.\\ 

{\small
\bibliographystyle{ieee_fullname}
\bibliography{egbib}

\begin{thebibliography}{10}\itemsep=-1pt

\bibitem{arivazhagan2003texture}
S Arivazhagan and L Ganesan.
\newblock Texture segmentation using wavelet transform.
\newblock {\em Pattern Recognition Letters}, 24(16):3197--3203, 2003.

\bibitem{deeplabv2}
Liang-Chieh Chen, George Papandreou, Iasonas Kokkinos, Kevin Murphy, and Alan~L
  Yuille.
\newblock Deeplab: Semantic image segmentation with deep convolutional nets,
  atrous convolution, and fully connected crfs.
\newblock {\em IEEE transactions on pattern analysis and machine intelligence},
  40(4):834--848, 2017.

\bibitem{deeplabv3}
Liang-Chieh Chen, George Papandreou, Florian Schroff, and Hartwig Adam.
\newblock Rethinking atrous convolution for semantic image segmentation.
\newblock {\em arXiv preprint arXiv:1706.05587}, 2017.

\bibitem{deeplabv3+}
Liang-Chieh Chen, Yukun Zhu, George Papandreou, Florian Schroff, and Hartwig
  Adam.
\newblock Encoder-decoder with atrous separable convolution for semantic image
  segmentation.
\newblock In {\em Proceedings of the European conference on computer vision
  (ECCV)}, pages 801--818, 2018.

\bibitem{reconet}
Wanli Chen, Xinge Zhu, Ruoqi Sun, Junjun He, Ruiyu Li, Xiaoyong Shen, and Bei
  Yu.
\newblock Tensor low-rank reconstruction for semantic segmentation.
\newblock In {\em European Conference on Computer Vision}, pages 52--69.
  Springer, 2020.

\bibitem{bfp}
Henghui Ding, Xudong Jiang, Ai~Qun Liu, Nadia~Magnenat Thalmann, and Gang Wang.
\newblock Boundary-aware feature propagation for scene segmentation.
\newblock In {\em Proceedings of the IEEE/CVF International Conference on
  Computer Vision (ICCV)}, October 2019.

\bibitem{svcnet}
Henghui Ding, Xudong Jiang, Bing Shuai, Ai~Qun Liu, and Gang Wang.
\newblock Semantic correlation promoted shape-variant context for segmentation.
\newblock In {\em Proceedings of the IEEE Conference on Computer Vision and
  Pattern Recognition}, pages 8885--8894, 2019.

\bibitem{danet}
Jun Fu, Jing Liu, Haijie Tian, Yong Li, Yongjun Bao, Zhiwei Fang, and Hanqing
  Lu.
\newblock Dual attention network for scene segmentation.
\newblock In {\em Proceedings of the IEEE Conference on Computer Vision and
  Pattern Recognition}, pages 3146--3154, 2019.

\bibitem{gonzales2002digital}
Rafael~C Gonzales and Richard~E Woods.
\newblock Digital image processing, 2002.

\bibitem{haralick1973textural}
Robert~M Haralick, Karthikeyan Shanmugam, and Its'~Hak Dinstein.
\newblock Textural features for image classification.
\newblock {\em IEEE Transactions on systems, man, and cybernetics},
  (6):610--621, 1973.

\bibitem{dmnet}
Junjun He, Zhongying Deng, and Yu Qiao.
\newblock Dynamic multi-scale filters for semantic segmentation.
\newblock In {\em Proceedings of the IEEE International Conference on Computer
  Vision}, pages 3562--3572, 2019.

\bibitem{hu2020class}
Hanzhe Hu, Deyi Ji, Weihao Gan, Shuai Bai, Wei Wu, and Junjie Yan.
\newblock Class-wise dynamic graph convolution for semantic segmentation.
\newblock {\em arXiv preprint arXiv:2007.09690}, 2020.

\bibitem{senet}
Jie Hu, Li Shen, and Gang Sun.
\newblock Squeeze-and-excitation networks.
\newblock In {\em Proceedings of the IEEE conference on computer vision and
  pattern recognition}, pages 7132--7141, 2018.

\bibitem{ccnet}
Zilong Huang, Xinggang Wang, Lichao Huang, Chang Huang, Yunchao Wei, and Wenyu
  Liu.
\newblock Ccnet: Criss-cross attention for semantic segmentation.
\newblock In {\em Proceedings of the IEEE International Conference on Computer
  Vision}, pages 603--612, 2019.

\bibitem{edge}
Xiangtai Li, Xia Li, Li Zhang, Guangliang Cheng, Jianping Shi, Zhouchen Lin,
  Shaohua Tan, and Yunhai Tong.
\newblock Improving semantic segmentation via decoupled body and edge
  supervision.
\newblock {\em arXiv preprint arXiv:2007.10035}, 2020.

\bibitem{spygr}
Xia Li, Yibo Yang, Qijie Zhao, Tiancheng Shen, Zhouchen Lin, and Hong Liu.
\newblock Spatial pyramid based graph reasoning for semantic segmentation.
\newblock In {\em Proceedings of the IEEE/CVF Conference on Computer Vision and
  Pattern Recognition}, pages 8950--8959, 2020.

\bibitem{sfnet}
Xiangtai Li, Ansheng You, Zhen Zhu, Houlong Zhao, Maoke Yang, Kuiyuan Yang,
  Shaohua Tan, and Yunhai Tong.
\newblock Semantic flow for fast and accurate scene parsing.
\newblock In {\em European Conference on Computer Vision}, pages 775--793.
  Springer, 2020.

\bibitem{msci}
Di Lin, Yuanfeng Ji, Dani Lischinski, Daniel Cohen-Or, and Hui Huang.
\newblock Multi-scale context intertwining for semantic segmentation.
\newblock In {\em Proceedings of the European Conference on Computer Vision
  (ECCV)}, pages 603--619, 2018.

\bibitem{refinenet}
Guosheng Lin, Anton Milan, Chunhua Shen, and Ian Reid.
\newblock Refinenet: Multi-path refinement networks for high-resolution
  semantic segmentation.
\newblock In {\em Proceedings of the IEEE conference on computer vision and
  pattern recognition}, pages 1925--1934, 2017.

\bibitem{linda2001shapiro}
G Linda.
\newblock Shapiro, george c. stockman. computer vision.
\newblock 2001.

\bibitem{cacnet}
Jianbo Liu, Junjun He, Yu Qiao, Jimmy~S Ren, and Hongsheng Li.
\newblock Learning to predict context-adaptive convolution for semantic
  segmentation.
\newblock In {\em European Conference on Computer Vision}, pages 769--786.
  Springer, 2020.

\bibitem{lgad}
Zhikang Liu and Lanyun Zhu.
\newblock Label-guided attention distillation for lane segmentation.
\newblock {\em Neurocomputing}, 2021.

\bibitem{fcn}
Jonathan Long, Evan Shelhamer, and Trevor Darrell.
\newblock Fully convolutional networks for semantic segmentation.
\newblock In {\em Proceedings of the IEEE conference on computer vision and
  pattern recognition}, pages 3431--3440, 2015.

\bibitem{ramola2020study}
Ayushman Ramola, Amit~Kumar Shakya, and Dai Van~Pham.
\newblock Study of statistical methods for texture analysis and their modern
  evolutions.
\newblock {\em Engineering Reports}, 2(4):e12149, 2020.

\bibitem{unet}
Olaf Ronneberger, Philipp Fischer, and Thomas Brox.
\newblock U-net: Convolutional networks for biomedical image segmentation.
\newblock In {\em International Conference on Medical image computing and
  computer-assisted intervention}, pages 234--241. Springer, 2015.

\bibitem{learnable}
Zhe Wang, Hongsheng Li, Wanli Ouyang, and Xiaogang Wang.
\newblock Learnable histogram: Statistical context features for deep neural
  networks.
\newblock In {\em European Conference on Computer Vision}, pages 246--262.
  Springer, 2016.

\bibitem{verydeep}
Zifeng Wu, Chunhua Shen, and Anton van~den Hengel.
\newblock Bridging category-level and instance-level semantic image
  segmentation.
\newblock {\em arXiv preprint arXiv:1605.06885}, 2016.

\bibitem{denseaspp}
Maoke Yang, Kun Yu, Chi Zhang, Zhiwei Li, and Kuiyuan Yang.
\newblock Denseaspp for semantic segmentation in street scenes.
\newblock In {\em Proceedings of the IEEE Conference on Computer Vision and
  Pattern Recognition}, pages 3684--3692, 2018.

\bibitem{cpnet}
Changqian Yu, Jingbo Wang, Changxin Gao, Gang Yu, Chunhua Shen, and Nong Sang.
\newblock Context prior for scene segmentation.
\newblock In {\em Proceedings of the IEEE/CVF Conference on Computer Vision and
  Pattern Recognition}, pages 12416--12425, 2020.

\bibitem{bisenet}
Changqian Yu, Jingbo Wang, Chao Peng, Changxin Gao, Gang Yu, and Nong Sang.
\newblock Bisenet: Bilateral segmentation network for real-time semantic
  segmentation.
\newblock In {\em Proceedings of the European Conference on Computer Vision
  (ECCV)}, pages 325--341, 2018.

\bibitem{ocr}
Yuhui Yuan, Xilin Chen, and Jingdong Wang.
\newblock Object-contextual representations for semantic segmentation.
\newblock {\em arXiv preprint arXiv:1909.11065}, 2019.

\bibitem{acfnet}
Fan Zhang, Yanqin Chen, Zhihang Li, Zhibin Hong, Jingtuo Liu, Feifei Ma, Junyu
  Han, and Errui Ding.
\newblock Acfnet: Attentional class feature network for semantic segmentation.
\newblock In {\em Proceedings of the IEEE International Conference on Computer
  Vision}, pages 6798--6807, 2019.

\bibitem{contextencoding}
Hang Zhang, Kristin Dana, Jianping Shi, Zhongyue Zhang, Xiaogang Wang, Ambrish
  Tyagi, and Amit Agrawal.
\newblock Context encoding for semantic segmentation.
\newblock In {\em Proceedings of the IEEE conference on Computer Vision and
  Pattern Recognition}, pages 7151--7160, 2018.

\bibitem{encnet}
Hang Zhang, Kristin Dana, Jianping Shi, Zhongyue Zhang, Xiaogang Wang, Ambrish
  Tyagi, and Amit Agrawal.
\newblock Context encoding for semantic segmentation.
\newblock In {\em Proceedings of the IEEE conference on Computer Vision and
  Pattern Recognition}, pages 7151--7160, 2018.

\bibitem{deepten}
Hang Zhang, Jia Xue, and Kristin Dana.
\newblock Deep ten: Texture encoding network.
\newblock In {\em Proceedings of the IEEE conference on computer vision and
  pattern recognition}, pages 708--717, 2017.

\bibitem{cfnet}
Hang Zhang, Han Zhang, Chenguang Wang, and Junyuan Xie.
\newblock Co-occurrent features in semantic segmentation.
\newblock In {\em Proceedings of the IEEE Conference on Computer Vision and
  Pattern Recognition}, pages 548--557, 2019.

\bibitem{pspnet}
Hengshuang Zhao, Jianping Shi, Xiaojuan Qi, Xiaogang Wang, and Jiaya Jia.
\newblock Pyramid scene parsing network.
\newblock In {\em Proceedings of the IEEE conference on computer vision and
  pattern recognition}, pages 2881--2890, 2017.

\bibitem{psanet}
Hengshuang Zhao, Yi Zhang, Shu Liu, Jianping Shi, Chen Change~Loy, Dahua Lin,
  and Jiaya Jia.
\newblock Psanet: Point-wise spatial attention network for scene parsing.
\newblock In {\em Proceedings of the European Conference on Computer Vision
  (ECCV)}, pages 267--283, 2018.

\bibitem{dgcwnet}
Lanyun Zhu, Shiping Zhu, Xuanyi Liu, and Li Luo.
\newblock Distance guided channel weighting for semantic segmentation.
\newblock {\em arXiv preprint arXiv:2004.12679}, 2020.

\bibitem{ann}
Zhen Zhu, Mengde Xu, Song Bai, Tengteng Huang, and Xiang Bai.
\newblock Asymmetric non-local neural networks for semantic segmentation.
\newblock In {\em Proceedings of the IEEE International Conference on Computer
  Vision}, pages 593--602, 2019.

\end{thebibliography}
}

\end{document}